\documentclass{article}

\usepackage[preprint]{neurips_2025}

\usepackage[utf8]{inputenc} % allow utf-8 input
\usepackage[T1]{fontenc}    % use 8-bit T1 fonts
\usepackage{hyperref}       % hyperlinks
\usepackage{url}            % simple URL typesetting
\usepackage{booktabs}       % professional-quality tables
\usepackage{amsfonts}       % blackboard math symbols
\usepackage{nicefrac}       % compact symbols for 1/2, etc.
\usepackage{microtype}      % microtypography
\usepackage{xcolor}         % colors
\usepackage{graphicx}
\usepackage{wrapfig}
\usepackage{amsmath}
\usepackage{tabularx}
\usepackage{subcaption}
\usepackage[table]{xcolor} % 必须在导言区添加此宏包
\usepackage{algorithm}
\usepackage{algpseudocode}

\definecolor{myred}{RGB}{247,241,234} % 定义一种深红
\definecolor{lightorange}{RGB}{250,239,239}

\title{FAAR: Format-Aware Adaptive Rounding for NVFP4}

\author{%
  Hanglin Li , 
  Shuchang Tian , 
  Chen Lin, 
  Zhiyong Zhao, 
  Kun Zhan
   \\
  Li Auto Inc. \\
  \texttt{\{lihanglin, tianshuchang, linchen, zhaozhiyong1, zhankun\}@lixiang.com} \\
}

\begin{document}

\maketitle

\begin{abstract}
Deploying large language models (LLMs) on edge devices requires extremely low-bit quantization. Ultra-low precision formats such as NVFP4 offer a promising solution for reducing memory footprint and accelerating computation. However, existing quantization methods typically rely on conventional rounding strategies and fail to account for the non-uniformity of the NVFP4 numerical grid, resulting in suboptimal rounding decisions and amplified quantization errors.
To address this, we propose \textbf{Format-Aware Adaptive Rounding (FAAR)}, a learnable rounding strategy tailored for the NVFP4 format. Unlike conventional quantization paradigms, FAAR explicitly incorporates the non-uniform NVFP4 grid into the optimization process. By adaptively adjusting rounding decisions guided by loss gradients, our method effectively approximates the theoretically optimal quantization. To complement FAAR, we introduce a \textbf{2-stages Format Alignment (2FA)} fine-tuning scheme that aligns LLM parameters layer-by-layer to the NVFP4 numerical space, further narrowing the performance gap. Remarkably, this learnable optimization incurs a minimal training overhead of only 4 GPU hours on Llama3-1B. Extensive experiments demonstrate the effectiveness of our approach. Compared with Round-to-Nearest (RTN), our method reduces perplexity on WikiText-2 from 14.28 to 12.60 on Llama3-1B and from 23.06 to 21.27 on Qwen3-1.7B. Additionally, our method consistently outperforms state-of-the-art approaches across various zero-shot downstream tasks.
\end{abstract}

\section{Introduction}
The substantial computational and memory demands of LLMs make edge device inference extremely challenging. To address this, low-bit quantization has emerged as a promising direction for reducing memory footprint and accelerating inference. Recently, NVIDIA introduced an ultra-low precision numerical format, NVFP4, designed to further reduce memory usage and improve computational efficiency~\cite{abecassis2025pretraining}. When deployed on Blackwell GPUs, NVFP4 can theoretically achieve a $4$--$6\times$ speedup compared to BF16 inference. Unlike conventional integer quantization formats, NVFP4 adopts a low-bit floating-point representation (E2M1) combined with block-wise scaling. This design significantly reduces storage requirements while maintaining a relatively wide numerical range, making it particularly attractive for ultra-low precision LLM deployment.

\begin{figure}[t] % [htbp] 控制浮动位置：here, top, bottom, page
  \centering
  \includegraphics[width=1.0\textwidth]{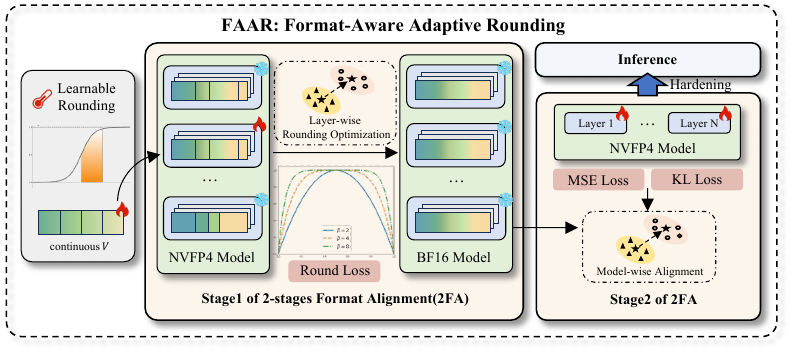} % 这里设置图片宽度为页面宽度的60%
  \caption{\textbf{The overall pipeline of our proposed quantization framework combining FAAR and 2FA.} 
\textbf{(1) FAAR + Stage 1:} 
The continuous learnable rounding variables $V$ are carefully initialized and incorporated into the layer-wise NVFP4 optimization. During this stage, we perform layer-wise rounding optimization against the frozen BF16 model. For each optimization step, only the current layer is updated while the rest of the network remains frozen.
The rounding decisions are optimized via learnable variables parameterized by a differentiable sigmoid function and guided by Round Loss.
\textbf{(2) FAAR + Stage 2:} 
The quantized layers are assembled into a full NVFP4 model and jointly optimized. Model-wise alignment is performed using KL divergence and last-hidden-state MSE losses to mitigate full-model error accumulation.
\textbf{(3) Hardening and Inference:} 
After the two-stage optimization, the continuous variables $V$ are deterministically hardened into discrete binary decisions, which are then deployed for efficient inference.}

  \label{pipline} % 给图片加标签，方便引用
\end{figure}

Despite these advantages, applying NVFP4 to LLMs remains non-trivial. A fundamental difficulty arises directly from the highly non-uniform numerical grid of NVFP4. Unlike uniform integer quantization where representable values are evenly spaced, NVFP4 defines a small set of irregularly distributed representable nodes. Most existing quantization techniques, including advanced approaches such as AdaRound~\cite{nagel2020up} and BRECQ~\cite{li2021brecq}, are designed for uniform quantization grids or rely on conventional rounding strategies such as round-to-nearest (RTN) or stochastic rounding (SR). However, these strategies implicitly assume uniform spacing between quantization levels and therefore fail to account for the irregular density of NVFP4 values. Because the gap between adjacent NVFP4 nodes widens significantly for larger magnitudes, a fixed rounding threshold (e.g., 0.5 in RTN) no longer reflects the optimal assignment for minimizing output error. As a result, rounding decisions become suboptimal, leading to amplified quantization errors and severe performance degradation in ultra-low precision LLMs.

To address this issue, we propose \textbf{Format-Aware Adaptive Rounding (FAAR)}, a novel rounding strategy specifically designed for NVFP4 quantization. Instead of using fixed heuristics such as RTN or SR, FAAR introduces learnable parameters that adaptively adjust rounding decisions guided by loss gradients. FAAR explicitly incorporates the non-uniform structure of the NVFP4 numerical grid into the optimization process, enabling the model to learn format-aware rounding behaviors that better minimize quantization error. While FAAR significantly improves the rounding process, discrepancies between the quantized model and its full-precision (BF16) counterpart still accumulate across layers. To further reduce this gap, we introduce a \textbf{2-stages Format Alignment (2FA)} fine-tuning scheme. In the first stage, we perform layer-wise calibration to learn optimal NVFP4 rounding offsets using FAAR. In the second stage, we conduct an end-to-end alignment with the full-precision (BF16) model to recover global model performance. By combining format-aware rounding with staged fine-tuning, our framework effectively mitigates quantization errors and substantially improves the robustness of ultra-low precision LLMs.

Our main contributions can be summarized as follows:
\begin{itemize}
    \item We propose \textbf{FAAR}, a first learning-based rounding strategy specifically tailored for the irregular NVFP4 format. Furthermore, we introduce a \textbf{2FA} fine-tuning scheme. By synergistically combining FAAR and 2FA, our framework effectively and continuously approximates the theoretically optimal quantization.
    \item Extensive experiments on the Llama3 and Qwen3 families validate the superiority of our method. We achieve state-of-the-art performance across various benchmarks (e.g., reducing WikiText-2 perplexity by 1.06 on Llama3-1B and 1.41 on Qwen3-1.7B compared to GPTQ+4/6). These significant improvements incur a minimal fine-tuning cost of only 4 GPU hours on 1B models, highlighting the practical value for edge deployment.
    \item We first demonstrate that existing training-free rounding methods are fundamentally suboptimal for the non-uniform characteristics of the NVFP4 numerical grid, establishing the necessity for a format-aware optimization approach.
\end{itemize}

\section{Related Work}
\subsection{Low-Bit LLM Quantization}
Low-bit quantization has been widely studied as an effective approach for enabling efficient LLM inference on edge devices. Existing techniques have been proposed to improve accuracy without extensive retraining, including advanced calibration strategies~\cite{lin2024awq,xiao2023smoothquant} and reconstruction-based optimization techniques~\cite{frantar2022gptq}. Furthermore, several works have explored aggressive low-bit schemes, such as 4-bit weight quantization~\cite{bongi2018olive,lin2025qserve,zhang2024qqq,shao2023omniquant,kim2023squeezellm,lee2024owq,dettmers2023spqr} and even binary networks~\cite{bai2021binarybert,qin2022bibert,liu2022bit}. Despite these advances, quantizing both weights and activations to extremely low precision remains a formidable challenge. While weight-only quantization can often maintain strong performance, fully quantized models typically suffer from severe accuracy degradation. This is largely caused by the highly dynamic and heavy-tailed activation distributions in LLMs~\cite{dettmers2022gpt3int8,lin2024duquant,guo2023olive,wei2023outlier}. Overcoming this W4A4 (4-bit weight and activation) performance drop necessitates more robust alignment and fine-tuning strategies.

\subsection{Emerging Quantization of NVFP4}
To further push the boundaries of hardware efficiency, recent research has shifted toward ultra-low precision floating-point formats. NVFP4, proposed by NVIDIA for Blackwell GPUs~\cite{abecassis2025pretraining}, has emerged as a highly promising standard. Unlike traditional integer formats, NVFP4 employs a non-uniform numerical grid combined with block-wise scaling, enabling a highly compact representation while preserving a wide dynamic range. Alongside NVFP4, related microscaling formats like MXFP4~\cite{ocp2023mx} have also been introduced to optimize deep learning workloads.
Despite the growing hardware support, algorithmic research on NVFP4 quantization is still in its infancy. Recent studies attempt to bridge this gap: MR-GPTQ~\cite{egiazarian2025bridging} adapts the GPTQ procedure to the numerical structure of low-bit floating-point formats; Four over Six~\cite{cook2025four} introduces an optimized NVFP4 scaling algorithm that significantly narrows the gap between low-precision and BF16 performance; TetraJet-v2~\cite{chen2025tetrajet} employs unbiased double-block quantization to mitigate outlier-induced inaccuracies; M2XFP~\cite{hu2026m2xfp} introduces an algorithm-hardware co-design for microscaling formats; and MS-EDEN~\cite{panferov2026quartet} proposes unbiased gradient estimation for end-to-end NVFP4 training. 

\subsection{Adaptive Rounding Strategies}
To minimize the information loss inherent in low-bit quantization, another line of research focuses on learnable rounding strategies. Recognizing that traditional RTN often yields suboptimal results, methods like AdaRound~\cite{nagel2020up} and BRECQ~\cite{li2021brecq} introduce learnable parameters to optimize rounding decisions. These adaptive techniques have demonstrated strong performance for uniform INT8 and INT4 settings. However, a fundamental limitation of these methods is their implicit assumption of uniform quantization grids, where the distance between adjacent levels is strictly constant. This assumption completely breaks down for floating-point-based ultra-low precision formats like NVFP4, which feature a non-uniform distribution of representable values. Consequently, directly applying conventional adaptive rounding formulations to these formats leads to inaccurate gradient estimation and suboptimal convergence.

While existing NVFP4 literature primarily focuses on scaling optimization or relies on fixed rounding heuristics, the optimization of discrete rounding decisions on irregular grids remains unexplored. To bridge this gap, our work extends the paradigm of adaptive rounding by proposing a format-aware framework that explicitly embeds the non-uniform NVFP4 representation into the rounding optimization process.

\section{Methodology}
\label{problem_nvfp4}
\subsection{NVFP4 Overview}
NVFP4 is an ultra-low precision floating-point format based on the FP4 (E2M1) representation combined with block-wise scaling. It defines a small set of non-uniformly distributed numerical nodes: $\mathcal{N} = \{0.0, \pm0.5, \pm1.0, \pm1.5, \pm2.0, \pm3.0, \pm4.0, \pm6.0\}$. To maintain a wide dynamic range under low precision, NVFP4 adopts a two-level scaling mechanism. First, the weight tensor $\mathbf{W}$ is partitioned into small blocks (e.g., 16 elements), and each block shares a local scaling factor $s_g$. These per-block scales are stored in FP8 (E4M3) precision, allowing the quantization process to adapt to local variations in value distribution. Second, a global scaling factor $s_{global}$ is introduced for the entire tensor. This global factor is represented in high precision (FP32) and serves as a scale-of-scales that normalizes all block-wise scaling factors. The global scaling helps maintain numerical stability and ensures that the FP8 block scales remain within their representable range. For a weight element $w$ belonging to block $g$, the normalized value is computed as 
$\tilde{w} = \frac{w}{s_g \cdot s_{global}}$. Quantization is applied to the magnitude while preserving the original sign of the weight. The normalized value is mapped to one of the representable NVFP4 nodes in the set $\mathcal{N}$. 
Let $w_{lower}$ and $w_{upper}$ denote the two adjacent nodes satisfying $w_{lower} \le |\tilde{w}| \le w_{upper}, \quad w_{lower}, w_{upper} \in \mathcal{N}$. Conventional quantization methods select one of these nodes using a deterministic rule such as RTN. The resulting quantized weight can be written as $w_q = \operatorname{sign}(w)\cdot \hat{q}\cdot s_g \cdot s_{global}$, where $\hat{q} \in \{w_{lower}, w_{upper}\}$ is the selected NVFP4 node. Due to the irregular spacing of $\mathcal{N}$, the optimal choice between $w_{lower}$ and $w_{upper}$ is no longer determined solely by distance.

\begin{figure}[t]
  \centering
  \begin{subfigure}[b]{0.49\textwidth}
    \centering
    \includegraphics[width=\textwidth]{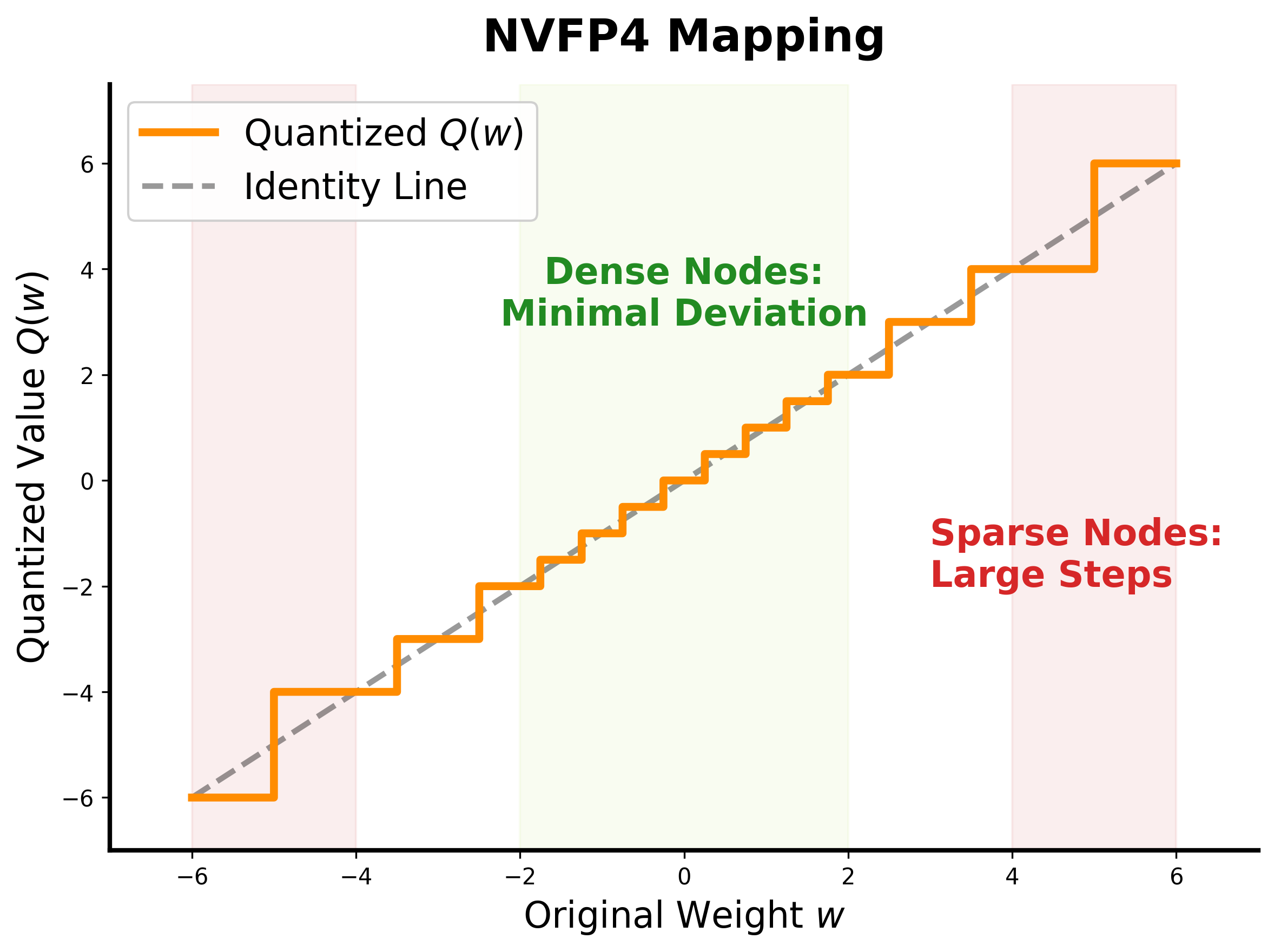} % 建议将原图裁剪为左右两张，或直接引用原图
    \caption{NVFP4 Mapping $Q(w)$}
    \label{fig:mapping-a}
  \end{subfigure}
  \hfill
  \begin{subfigure}[b]{0.49\textwidth}
    \centering
    \includegraphics[width=\textwidth]{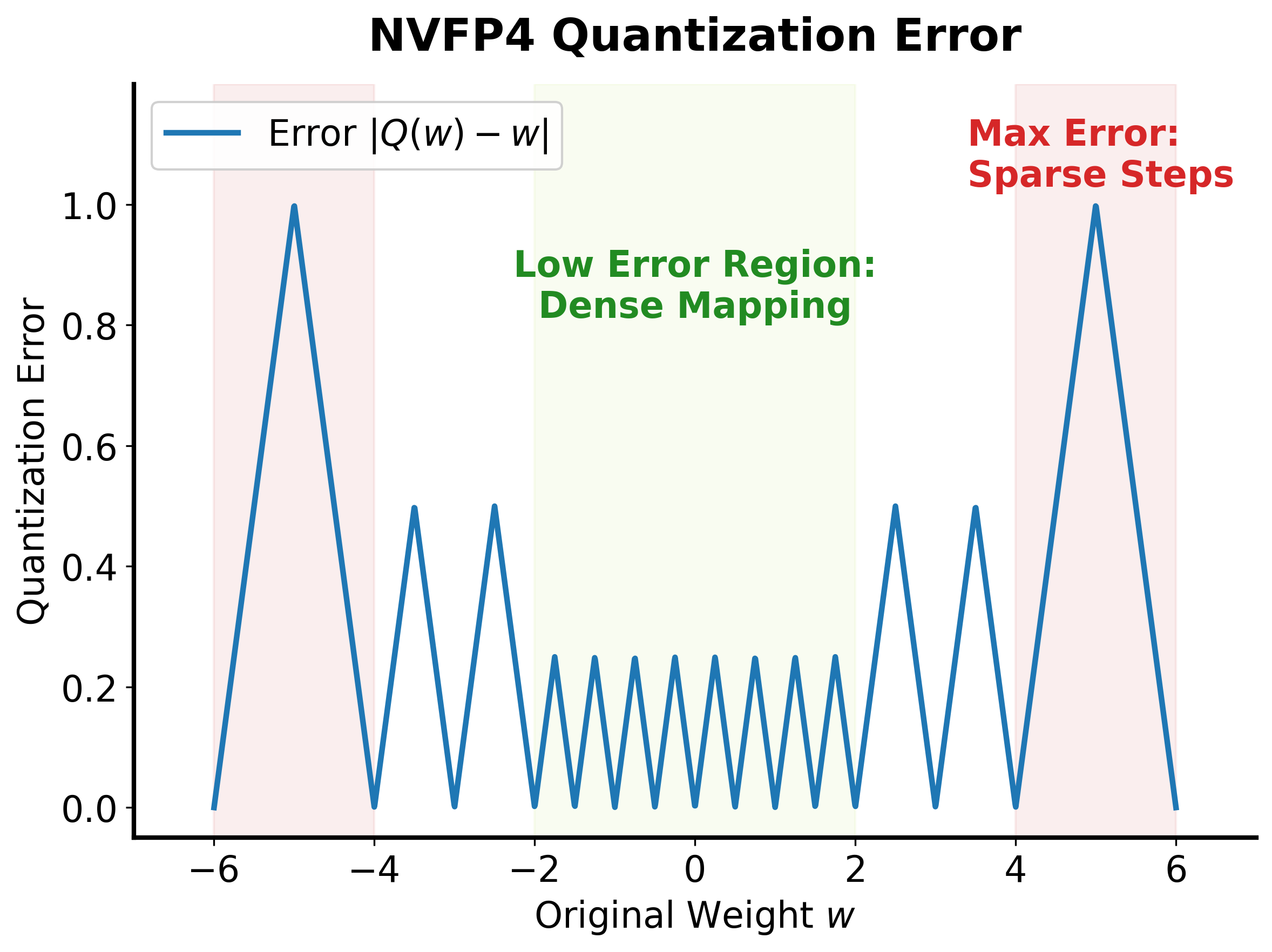}
    \caption{Quantization Error $|Q(w) - w|$}
    \label{fig:error-b}
  \end{subfigure}
  
  \caption{\textbf{The non-uniform NVFP4 grid introduces magnitude-dependent rounding errors.} 
  \textbf{(a)} The mapping function relative to the original weight $w$, showing numerical nodes that are densely concentrated near zero but become increasingly sparse for larger magnitudes. 
  \textbf{(b)} The absolute quantization error, highlighting the low-error region (shaded green) optimized for the core weight distribution and the amplified distortion for weights with larger magnitudes.}
  \label{quant-err}
\end{figure}

\subsection{Rounding Challenge of the Non-uniform NVFP4 Grid} 
The core challenge of NVFP4 quantization lies in the fact that rounding decisions are no longer locally optimal under a non-uniform grid. Unlike uniform quantization schemes, the NVFP4 grid exhibits an irregular distribution. As illustrated in Figure~\ref{quant-err}, the numerical nodes are densely concentrated near zero but become increasingly sparse for larger magnitudes. As a result, mapping continuous model weights to the NVFP4 grid introduces magnitude-dependent rounding errors. In particular, weights with larger magnitudes fall into wider quantization intervals, leading to amplified distortion that propagates through matrix multiplications across layers. Consequently, rounding errors often become a dominant source of performance degradation in NVFP4 quantization. Prior studies on neural network quantization have shown that the commonly used RTN strategy is not always optimal~\cite{li2021brecq,nagel2020up}. To examine whether this phenomenon persists in NVFP4 quantization, we conduct an empirical study on stochastic rounding strategies. Specifically, we sample 100 stochastic rounding candidates across all layers of Llama3-1B and evaluate the resulting perplexity (PPL). 
\begin{wraptable}{r}{0.5\textwidth} % {r} 表示靠右，0.5\textwidth 是占用的宽度，可根据需要调整
  \centering
  % \vspace{-10pt} % 可选：微调表格上方的间距
  \caption{\textbf{RTN is Suboptimal for NVFP4 Quantization.} We compare WikiText-2 perplexity across different rounding schemes for Llama3-1B.}
  \begin{tabular}{lr}
    \toprule
    Rounding scheme & PPL \\
    \midrule
    baseline & 13.04 \\
    lower & 13.05 \\
    upper & 13.08 \\
    \midrule
    Stochastic & $13.056 \pm 0.034$ \\
    Stochastic (best) & 13.02 \\
    \bottomrule
  \end{tabular}
  \vspace{-10pt} % 可选：微调表格上方的间距
  \label{rounding_comparison}
\end{wraptable}
As shown in Table~\ref{rounding_comparison}, both deterministic lower and upper rounding underperform the RTN baseline, with upper rounding exhibiting a 0.04 increase in PPL compared to RTN. However, among only 100 stochastic rounding trials, 13 configurations achieve lower PPL than RTN, with the best one yielding a PPL reduction of over 0.02. These results indicate that, under a non-uniform grid, the optimal rounding assignment cannot be recovered by any local or heuristic rule, and instead requires explicit optimization. The existence of better rounding assignments, even under such a limited random search budget, strongly motivates our learnable rounding approach.

\subsection{Problem Definition}
\label{Definition}
Our empirical observations suggest that selecting appropriate rounding assignments is critical for minimizing quantization error under the NVFP4 format, and superior rounding solutions exist within the search space. Formally, let $N$ denote the number of quantized weights, and $\mathbf{V} \in \{0,1\}^N$ denote the rounding decision tensor, where each element determines whether to round to $w_{lower}$ or $w_{upper}$. Our goal is to find the optimal rounding assignment $\mathbf{V}^*$ that minimizes the task loss $\mathcal{L}$:
\begin{equation}
\mathbf{V}^* = \mathop{\arg\min}_{\mathbf{V}} \mathcal{L}\big(\mathbf{W}_q(\mathbf{V})\big).
\label{eq1}
\end{equation}
However, searching for $\mathbf{V}^*$ over the discrete space is computationally intractable. To address this, we propose FAAR, a learnable rounding strategy that optimizes rounding decisions based on the local NVFP4 node context of each weight element. Furthermore, we introduce a 2FA fine-tuning scheme that progressively aligns LLM parameters with the NVFP4 format and further reduces the performance gap.

\subsection{Format-Aware Adaptive Rounding}
As formulated in Section~\ref{Definition}, directly searching for the optimal discrete rounding assignment $\mathbf{V}^*$ is computationally intractable. To address this, we propose FAAR to relax the discrete rounding decisions into a continuous optimization problem, enabling gradient-based optimization of rounding decisions. Unlike AdaRound~\cite{nagel2020up} designed for uniform integer quantization grids, FAAR ensures that the optimization is sensitive to the varying interval widths, allowing larger-interval weights to receive stronger corrective gradients.

\paragraph{Continuous Relaxation of NVFP4 Quantization}
Given a floating-point weight tensor $\mathbf{W}$, its group-wise scaling factors $s_g$, and the global scaling factor $s_{global}$, we first apply the two-level scaling mechanism. Since the NVFP4 format represents symmetric positive and negative values, we perform quantization on the magnitude while preserving the original sign of the weight. For each weight element $w \in \mathbf{W}$ in group $g$, we locate its two adjacent NVFP4 representable nodes $w_{lower}$ and $w_{upper}$ from the node set $\mathcal{N}$ such that $w_{lower} \le \frac{|w|}{s_g \cdot s_{global}} \le w_{upper}$. Instead of applying a hard, deterministic rounding rule, we introduce a learnable continuous variable $v \in [0,1]$ for each weight element. The quantized weight element $w_q$ is thus parameterized as:
\begin{equation}
w_{q} = \operatorname{sign}(w) \cdot
\Big[ w_{lower} + h_{\beta}(v) \cdot (w_{upper} - w_{lower}) \Big]
\cdot (s_g \cdot s_{global}),
\label{eq2}
\end{equation}
where $h_{\beta}(v)$ is a differentiable rounding function that maps the continuous variable $v$ to an interpolation coefficient. In uniform quantization, the distance between adjacent quantization nodes remains constant. In contrast, the NVFP4 node spacing $(w_{upper} - w_{lower})$ varies dynamically depending on the specific interval where $w$ resides. By explicitly incorporating this local interval span into the parameterization, FAAR adaptively scales the gradients for each rounding variable $v$, naturally accommodating the non-uniform quantization grid during optimization. Let $\mathbf{V}$ denote the tensor of learnable rounding variables $v$, having the same shape as $\mathbf{W}$. During optimization, $\mathbf{V}$ is updated via gradient descent.

\paragraph{Differentiable Soft Rounding}
To enable stable gradient-based optimization while gradually enforcing discrete decisions, we adopt a temperature-scaled sigmoid function for $h_{\beta}(v)$:
\begin{equation}
h_{\beta}(v) = \sigma\big(\beta(v-0.5)\big)
= \frac{1}{1 + e^{-\beta(v-0.5)}} .
\label{eq3}
\end{equation}
The temperature parameter $\beta$ controls the steepness of the function. During the initial phase of optimization, a relatively small $\beta$ provides a smooth mapping, facilitating stable gradient propagation. As training progresses, $\beta$ is gradually increased following an annealing schedule. In the limit of $\beta \to \infty$, $h_{\beta}(v)$ approaches a binary step function, effectively converting the soft rounding decision into a hard rounding operation.
\begin{table}[t]
\centering
\caption{
\textbf{Procedure for the FAAR and 2FA quantization framework.}
}
\label{tab:faar_tsfa}
\begin{tabular}{c l}
\toprule
\textbf{Step} & \textbf{Pseudocode} \\
\midrule

1 & \textbf{for each layer $l$ do} \\

2 & \quad $\tilde{\mathbf{W}} = |\mathbf{W}| / (s_g \cdot s_{global})$ \\

3 & \quad $(\mathbf{W}_{lower}, \mathbf{W}_{upper}) = \text{FindInterval}(\tilde{\mathbf{W}}, \mathcal{N})$ \\

4 & \quad $\mathbf{V}_{init} = (\tilde{\mathbf{W}}-\mathbf{W}_{lower}) / (\mathbf{W}_{upper}-\mathbf{W}_{lower})$ \\

5 & \quad Initialize learnable variable $\mathbf{V} = \mathbf{V}_{init}$ \\

6 & \quad $h_{\beta}(\mathbf{V}) = \sigma\big(\beta(\mathbf{V}-0.5)\big)$ \\

7 & \quad $\mathbf{W}_q =
\operatorname{sign}(\mathbf{W})
\odot
\big[\mathbf{W}_{lower} + h_{\beta}(\mathbf{V}) \odot (\mathbf{W}_{upper}-\mathbf{W}_{lower})\big]
\odot (s_g \cdot s_{global})$ \\

8 & \quad $\mathbf{Y}_{fp} = \mathbf{X}\mathbf{W}$ \\

9 & \quad $\mathbf{Y}_q = \mathbf{X}_q\mathbf{W}_q$ \\

10 & \quad $\mathcal{L}_{MSE} = \|\mathbf{Y}_{fp}-\mathbf{Y}_q\|_F^2$ \\

11 & \quad $\mathcal{L}_{round} = \frac{1}{N}\sum_{i=1}^{N} \big(1-(2v_i-1)^2\big)$ \\

12 & \quad $\mathcal{L}_{stage1} =  \mathcal{L}_{MSE} + \lambda_{round}\mathcal{L}_{round}$ \\

13 & \quad Update $\mathbf{V}$ by minimizing $\mathcal{L}_{stage1}$ \\

14 & \textbf{end for} \\

\midrule

15 & Construct full NVFP4 model parameterized by rounding variables $\mathbf{V}$ \\

16 & \textbf{for training steps do} \\

17 & \quad Compute logits $\mathbf{Z}_{fp}, \mathbf{Z}_q$ and last hidden states $\mathbf{H}_{fp}, \mathbf{H}_q$ \\

18 & \quad $\mathbf{P}_{fp} = \text{softmax}(\mathbf{Z}_{fp} / \tau), \mathbf{P}_q = \text{softmax}(\mathbf{Z}_q / \tau)$ \\

19 & \quad $\mathcal{L}_{KL} = \mathrm{KL}(\mathbf{P}_{fp} \| \mathbf{P}_q)$\\

20 & \quad $\mathcal{L}_{MSE} = \|\mathbf{H}_{fp}-\mathbf{H}_q\|_F^2$ \\

21 & \quad $\mathcal{L}_{round} = \sum_{l=1}^{L} \mathcal{L}_{round}^{(l)}$ \\

22 & \quad $\mathcal{L}_{stage2} =
\lambda_{KL}\mathcal{L}_{KL}
+
\mathcal{L}_{MSE}
+
\lambda_{round}\mathcal{L}_{round}$ \\

23 & \quad Update rounding variables $\mathbf{V}$ by minimizing $\mathcal{L}_{stage2}$ \\

24 & \textbf{end for} \\

\midrule

25 & $\mathbf{V}_{harden} =
\begin{cases}
1 & \mathbf{V} \ge 0.5 \\
0 & \text{otherwise}
\end{cases}$ \\

26 & $\mathbf{W}_{final} =
\operatorname{sign}(\mathbf{W})
\odot
\big[\mathbf{W}_{lower}+\mathbf{V}_{harden} \odot (\mathbf{W}_{upper}-\mathbf{W}_{lower})\big]
\odot (s_g \cdot s_{global})$ \\

\bottomrule
\end{tabular}
\end{table}

\paragraph{Initialization of Rounding Variables}
To accelerate convergence and avoid poor local minima, we initialize the learnable variable $v$ based on the exact relative position of each weight within its NVFP4 interval:
\begin{equation}
v_{init} = \frac{\frac{|w|}{s_g \cdot s_{global}} - w_{lower}}{w_{upper} - w_{lower}}.
\label{eq4}
\end{equation}
This initialization preserves the relative position of the floating-point value within its NVFP4 interval before any rounding occurs, placing the variables close to their expected optimal values. Thus, we set $v = v_{init}$ at the beginning of the optimization.

\subsection{Two-Stage Format Alignment Training}
While FAAR effectively minimizes local quantization errors, optimizing the rounding variables of each layer independently ignores inter-layer error propagation. Quantization errors introduced in early layers accumulate and alter the input activation distributions of later layers, leading to a mismatch between local optimization targets and global model accuracy. To mitigate this error accumulation and achieve optimal overall performance, we propose a progressive Two-Stage Format Alignment Training strategy, which transitions from local layer-wise rounding to global full-model fine-tuning.

\paragraph{Stage 1: Layer-wise Adaptive Rounding}
In the first stage, we optimize the rounding variables $\mathbf{V}$ for each layer independently. For a given layer, the input activations $\mathbf{X}$ are sampled from the pre-trained full-precision (BF16) model, and $\mathbf{X}_q$ denotes the quantized activations. To isolate the optimization process and find high-quality local rounding decisions, we minimize the discrepancy between the exact floating-point output $\mathbf{X}\mathbf{W}$ and the quantized output $\mathbf{X}_q\mathbf{W}_q(\mathbf{V})$:
\begin{equation}
\mathcal{L}_{stage1} = \big\| \mathbf{X}\mathbf{W} - \mathbf{X}_q\mathbf{W}_q(\mathbf{V}) \big\|_F^2 + \lambda_{round}\frac{1}{N} \sum_{i=1}^{N} \Big(1 - (2v_i - 1)^2\Big).
\label{eq5}
\end{equation}
The first term is the Frobenius norm of the output reconstruction error. The second term acts as a regularization penalty that pushes the continuous variables $v_i$ towards the binary values $\{0, 1\}$, thereby reducing the discrepancy between soft and hard rounding. To ensure numerical stability and valid regularization, the learnable variable $v_i$ is clipped to the interval $[0, 1]$ after each gradient update. The optimized rounding variables obtained in this stage serve as a robust initialization for the subsequent global alignment.

\paragraph{Stage 2: Full-Model Alignment}
After obtaining the locally optimized rounding variables for all layers, we assemble the full NVFP4 quantized model. To account for inter-layer error propagation, we perform a global alignment against the original full-precision (BF16) model. Given the outputs and hidden representations produced by both models, we optimize the following joint objective:
\begin{equation}
\mathcal{L}_{stage2} =
\lambda_{KL}\mathcal{L}_{KL}
+
\mathcal{L}_{MSE}
+
\lambda_{round}\sum_{l=1}^{L} \mathcal{L}_{round}^{(l)},
\label{eq6}
\end{equation}
where $\mathcal{L}_{KL}$ measures the Kullback-Leibler divergence between the final output probability distributions of the two models, $\mathcal{L}_{MSE}$ aligns the last hidden representations of the two models, $L$ denotes the number of quantized layers, and $\mathcal{L}_{round}^{(l)}$ is the rounding regularization loss of the $l$-th layer.

\paragraph{Hardening and Inference}
Upon convergence of the two-stage training, the continuous variables are deterministically hardened into discrete binary decisions:
\begin{equation}
\hat{v} = \mathbb{I}\big(v \ge 0.5\big) \in \{0, 1\},
\label{eq7}
\end{equation}
where $\mathbb{I}(\cdot)$ is the indicator function. The final NVFP4 weights are obtained by substituting $h_{\beta}(v)$ with $\hat{v}$ in Eq.~(\ref{eq2}). The resulting weights follow the NVFP4 numerical format and are directly deployable on NVFP4 hardware.

\section{Evaluation}
\subsection{Experimental Settings}
\label{Settings}
\paragraph{Models and Tasks} 
To demonstrate the effectiveness of our proposed quantization method, we evaluate its performance across various LLMs, specifically the Llama3~\cite{grattafiori2024llama3} and Qwen3~\cite{yang2025qwen3} families. We compare our approach against several state-of-the-art PTQ baselines, including RTN, GPTQ~\cite{frantar2022gptq}, MR-GPTQ~\cite{egiazarian2025bridging}, the recently proposed 4/6 method~\cite{cook2025four}, and the combined GPTQ+4/6. The evaluation is conducted on standard WikiText-2 and C4 benchmarks. We report PPL to measure language generation capability and cosine similarity to assess the preservation of the internal activation space. Furthermore, we evaluate our method across downstream tasks including BoolQ~\cite{clark2019boolq}, Arc-Easy, Arc-Challenge~\cite{clark2018arc}, and HellaSwag~\cite{zellers2019hellaswag}. All downstream evaluations are conducted in a zero-shot setting using the LM Evaluation Harness. Experiments are conducted on a single NVIDIA H200 GPU, where the Llama3-1B model requires approximately 4 GPU hours.

\paragraph{Baseline} To analyze the contribution of our proposed optimization strategy, we propose an intermediate variant denoted as a strong baseline. This strong baseline is constructed by enhancing the conventional RTN quantization with several practical improvements, serving as the foundation for all subsequent experiments. As shown in Table~\ref{ppl_results_full}, the strong baseline provides a more stable starting point than the naive RTN quantization. The benefit of this additional optimization over the strong baseline is clearly reflected in the perplexity results. For example, on Llama3-1B evaluated on WikiText-2, the perplexity decreases from 14.03 with the strong baseline to 12.60 after applying the full optimization pipeline. Similar improvements are consistently observed across all tested model sizes and datasets. 

\subsection{Main Results}
\begin{table}[t]
\centering
\caption{
\textbf{FAAR Framework Consistently Outperforms Existing PTQ approaches.} Our method achieves the lowest word perplexity ($\downarrow$) across Llama3 and Qwen3 models on both WikiText-2 and C4 datasets.}
\label{ppl_results_full}
\begin{tabular}{lcccc|cccc}
\toprule
 & \multicolumn{4}{c}{WikiText-2 Word PPL ($\downarrow$)} & \multicolumn{4}{c}{C4 Word PPL ($\downarrow$)} \\
\cmidrule(lr){2-5} \cmidrule(lr){6-9}
 & \multicolumn{2}{c}{Llama3} & \multicolumn{2}{c}{Qwen3} & \multicolumn{2}{c}{Llama3} & \multicolumn{2}{c}{Qwen3} \\
\cmidrule(lr){2-3} \cmidrule(lr){4-5} \cmidrule(lr){6-7} \cmidrule(lr){8-9}
Method & 1B & 8B & 1.7B & 8B & 1B & 8B & 1.7B & 8B \\
\midrule
BF16 & 11.98 & 7.54 & 21.04 & 12.21 & 28.55 & 18.08 & 58.75 & 36.33 \\
\midrule
RTN & 14.28 & 8.44 & 23.06 & 12.67 & 36.19 & 20.82 & 65.54 & 37.91 \\
GPTQ & 13.74 & 8.35 & 21.48 & 12.49 & 35.62 & 20.96 & 63.34 & 37.14 \\
MR-GPTQ & 13.73 & 8.32 & 21.42 & 12.44 & 35.59 & 20.92 & 63.31 & 37.19 \\
4/6 & 13.89 & 8.30 & 23.57 & 12.55 & 35.10 & 20.45 & 66.30 & 37.31 \\
GPTQ+4/6 & 13.66 & 8.29 & 22.68 & 12.62 & 35.56 & 20.89 & 63.81 & 37.25 \\
\rowcolor{lightorange} Ours (strong baseline) & 14.03 & 8.40 & 22.89 & 12.62 & 35.82 & 20.73 & 65.41 & 37.64 \\
\rowcolor{lightorange} \textbf{Ours (FAAR+2FA)} & \textbf{12.60} & \textbf{8.13} & \textbf{21.27} & \textbf{12.32} & \textbf{34.41} & \textbf{20.17} & \textbf{62.24} & \textbf{36.92} \\
\bottomrule
\end{tabular}
\end{table}

\begin{table}[t]
\centering
\caption{
\textbf{FAAR Preserves the Internal Feature Space of BF16.} Our method yields higher cosine similarity ($\uparrow$) between quantized and BF16 last hidden states, indicating improved preservation of internal representations. Results are reported as percentages (\%).}

\label{cosine_results}
\begin{tabular}{lcccc|cccc}
\toprule
 & \multicolumn{4}{c}{WikiText-2 Cosine Similarity ($\uparrow$)} & \multicolumn{4}{c}{C4 Cosine Similarity ($\uparrow$)} \\
\cmidrule(lr){2-5} \cmidrule(lr){6-9}
 & \multicolumn{2}{c}{Llama3} & \multicolumn{2}{c}{Qwen3} & \multicolumn{2}{c}{Llama3} & \multicolumn{2}{c}{Qwen3} \\
\cmidrule(lr){2-3} \cmidrule(lr){4-5} \cmidrule(lr){6-7} \cmidrule(lr){8-9}
Method & 1B & 8B & 1.7B & 8B & 1B & 8B & 1.7B & 8B \\
\midrule
BF16 & 100.00 & 100.00 & 100.00 & 100.00 & 100.00 & 100.00 & 100.00 & 100.00 \\
\midrule
RTN & 96.08 & 97.23 & 97.97 & 97.92 & 94.27 & 98.03 & 97.11 & 97.11 \\
GPTQ & 97.78 & 98.45 & 98.75 & 99.01 & 96.01 & 96.88 & 97.95 & 98.63 \\
MR-GPTQ & 97.90 & 97.80 & 98.76 & 99.08 & 97.02 & 97.41 & 98.02 & 98.47 \\
4/6 & 96.72 & 98.50 & 96.94 & 98.92 & 97.33 & 98.28 & 95.61 & 97.91 \\
GPTQ+4/6 & 97.93 & 98.62 & 98.68 & 98.49 & 97.14 & 97.55 & 97.34 & 98.42 \\
\rowcolor{lightorange} Ours (strong baseline) & 96.55 & 97.78 & 98.22 & 98.59 & 95.87 & 98.11 & 97.21 & 97.82 \\
\rowcolor{lightorange} \textbf{Ours (FAAR+2FA)} & \textbf{98.06} & \textbf{98.96} & \textbf{99.02} & \textbf{99.13} & \textbf{97.50} & \textbf{98.39} & \textbf{98.21} & \textbf{98.71} \\
\bottomrule
\end{tabular}
\end{table}
\paragraph{Language Modeling Performance} The PPL results for the evaluated models under different quantization settings are summarized in Table~\ref{ppl_results_full}. Lower perplexity indicates better preservation of the original language modeling capabilities of the model.

As shown in Table~\ref{ppl_results_full}, the naive RTN strategy suffers from severe performance degradation. While methods incorporating second-order Hessian information (e.g., GPTQ and MR-GPTQ) effectively mitigate quantization errors, our method consistently outperforms them across all model sizes and datasets. For instance, on Llama3-1B (WikiText-2), our method achieves a PPL of 12.60. This significantly outperforms RTN (14.28) and GPTQ (13.74), and closely approaches the full-precision BF16 baseline (11.98). The 4/6 method improves the dynamic range for numbers quantized between 4 and 6, and its combination with GPTQ (GPTQ+4/6) serves as a strong baseline. However, our method still demonstrates clear superiority. On the C4 benchmark for Qwen3-1.7B, our method achieves a PPL of 62.24, which is a substantial improvement over both RTN baseline (65.54) and the strong GPTQ+4/6 baseline (63.81). This suggests that our format-aware adaptive rounding strategy is more effective at minimizing magnitude-dependent distortion than simply altering the numerical grid or applying static second-order penalties.

\paragraph{Feature Space Preservation} To further understand the source of performance improvements, we measure the cosine similarity between the last hidden states of the quantized models and their full-precision (BF16) counterparts. The results are presented in Table~\ref{cosine_results}. Higher similarity indicates that the quantized model faithfully preserves the original feature space.

Our method achieves the highest cosine similarity across all tested models and datasets, consistently exceeding 97.5\% and reaching up to 99.13\% (Qwen3-8B on WikiText-2). In contrast, the RTN baseline struggles to maintain feature alignment, dropping to 94.27\% on Llama3-1B (C4). Even advanced baselines like GPTQ+4/6 plateau around 97\%--98\%. The superior cosine similarity scores provide a strong explanation for the perplexity improvements observed in Table~\ref{ppl_results_full}. By explicitly optimizing rounding decisions based on the local NVFP4 quantization interval and aligning the outputs globally, our method effectively suppresses the compounding of quantization errors across layers. This high fidelity is maintained consistently across both Llama3 and Qwen3 architectures, demonstrating that our approach exhibits strong robustness and scalability for deploying LLMs under ultra-low precision constraints.

\paragraph{Downstream Task Performance}
While perplexity is a reliable indicator of language modeling quality, it does not always fully reflect performance on real-world tasks. Therefore, we further evaluate quantized models on a set of representative downstream benchmarks, including BoolQ~\cite{clark2019boolq}, Arc-Easy, Arc-Challenge~\cite{clark2018arc}, and HellaSwag. Following prior work, we report normalized accuracy to mitigate discrepancies caused by tokenization differences across models.

The results on Llama3-1B and Llama3-8B models are summarized in Table~\ref{downstream tasks}. Compared with the full-precision BF16 baseline, all quantization methods inevitably introduce performance degradation. However, our method consistently achieves the best trade-off between compression and accuracy. Specifically, on the 1B model, our method attains an average accuracy of 55.97, outperforming RTN by +3.44 points and GPTQ by +2.94 points, while also surpassing more advanced variants such as MR-GPTQ and GPTQ+4/6. Notably, our approach achieves performance close to the BF16 baseline (56.70), with only a marginal gap of 0.73 points, indicating that the proposed method effectively preserves task-level capabilities under extreme quantization. On the larger 8B model, a similar trend is observed. Our method achieves the highest average accuracy of 73.28, improving over RTN (+1.30) and GPTQ (+0.93), and maintaining consistent gains across all evaluated tasks. In particular, we observe clear improvements on more challenging benchmarks such as Arc-Challenge and HellaSwag, suggesting that our method better preserves complex reasoning and commonsense understanding abilities compared to prior approaches.

\begin{table}[t]
\centering
\caption{
\textbf{Downstream task performance of Llama3 models across various PTQ methods.} Results are reported as percentages (\%). }
% We evaluate the Llama3 model on common benchmarks including BoolQ, Arc-E, Arc-C, and HellaSwag.}
\label{downstream tasks}
\resizebox{\linewidth}{!}{
\begin{tabular}{lcccccccccc}
\toprule
 & \multicolumn{2}{c}{BoolQ($\uparrow$)} & \multicolumn{2}{c}{Arc-E($\uparrow$)} & \multicolumn{2}{c}{Arc-C($\uparrow$)} & \multicolumn{2}{c}{HellaSwag($\uparrow$)} & \multicolumn{2}{c}{\textbf{Average($\uparrow$)}} \\
\cmidrule(lr){2-3} \cmidrule(lr){4-5} \cmidrule(lr){6-7} \cmidrule(lr){8-9} \cmidrule(lr){10-11}
Method & 1B & 8B & 1B & 8B & 1B & 8B & 1B & 8B & 1B & 8B \\
\midrule
BF16 & 63.61 & 83.12 & 62.08 & 82.71 & 36.86 & 54.96 & 64.24 & 79.27 & 56.70 & 75.02 \\
\midrule
RTN  & 58.55 & 80.24 & 57.31 & 77.51 & 34.28 & 52.49 & 59.97 & 77.66 & 52.53 & 71.98 \\
MR-GPTQ & 61.27 & 80.89 & 57.38 & 78.55 & 33.47 & 53.70 & 60.65 & 77.20 & 53.19 & 72.59 \\
GPTQ & 61.06 & 80.28 & 57.41 & 78.30 & 33.07 & 53.81 & 60.58 & 77.02 & 53.03 & 72.35 \\
GPTQ+4/6 & 60.17 & 81.40 & 57.54 & 78.72 & 33.81 & 53.16 & 60.73 & 77.52 & 53.06 & 72.70 \\
\rowcolor{lightorange} \textbf{Ours (FAAR+2FA)} & \textbf{63.27} & \textbf{81.79} & \textbf{61.70} & \textbf{79.02} & \textbf{36.09} & \textbf{53.95} & \textbf{62.80} & \textbf{78.36} & \textbf{55.97} & \textbf{73.28} \\
\bottomrule
\end{tabular}
}
\end{table}

\subsection{Ablation Studies}
To isolate and quantify the contributions of our method, we conduct comprehensive ablation studies focusing on the efficacy of individual algorithmic components and the sensitivity of the model to key hyperparameters. Unless otherwise specified, all experiments adhere to the calibration setup detailed in Section~\ref{Settings}. 
\paragraph{Effect of Algorithmic Components} 
Our proposed quantization pipeline introduces two critical enhancements: FAAR and 2FA. To evaluate their respective impacts, we incrementally integrate these modules and measure the PPL on the WikiText-2 dataset.

As shown in Table~\ref{tab:ablation}, the progressive integration of our modules yields consistent perplexity reductions across both architectures. The introduction of FAAR provides a substantial performance uplift compared to the RTN baseline, reducing PPL from 14.28 to 13.01 on Llama3-1B. This demonstrates that explicitly optimizing continuous rounding variables over the non-uniform NVFP4 grid effectively compensates for the magnitude-dependent information loss inherent in standard quantization. Building upon this, the 2FA global alignment further refines the weights. By aligning the full-model output distributions and hidden states, this global alignment effectively mitigates inter-layer error accumulation. Consequently, the full method achieves the best perplexity (12.60 on Llama3-1B and 21.27 on Qwen3-1.7B), indicating that the local rounding optimization and global format alignment are highly complementary.

\begin{table}[t]
  \centering
  \caption{
  % Both FAAR and TSFA are Critical for Optimal Performance.
  \textbf{Effect of Algorithmic Components.} FAAR significantly reduces perplexity compared to RTN, and the addition of 2FA further mitigates errors and yields the best results (WikiText-2 Word PPL $\downarrow$).}
  \label{tab:ablation}
  \begin{tabular}{lcc}
    \toprule
    Method & Llama3-1B & Qwen3-1.7B \\
    \midrule
    BF16 & 11.98 & 21.04 \\
    \midrule
    RTN & 14.28 & 23.06 \\
    FAAR  & 13.01 & 21.86 \\
    \rowcolor{lightorange} \textbf{FAAR + 2FA} & \textbf{12.60} & \textbf{21.27} \\
    \bottomrule
  \end{tabular}
\end{table}

\begin{table}[t]
  \centering
  \begin{minipage}{0.46 \textwidth}
    \centering
    \caption{
    \textbf{Effect of Optimization Steps.} Increasing the number of steps reduces PPL ($\downarrow$), with diminishing returns beyond 2,500 steps. We adopt 2,500 steps (\underline{underlined}) as the default, achieving near-optimal performance with reduced computational cost.}
    \label{tab:steps-tuning}
    \begin{tabular}{ccc}
      \toprule
        Steps & Llama3-1B & Qwen3-1.7B \\
      \midrule
      0 & 13.01 & 21.86 \\
      500 & 12.84 & 21.52 \\
      \underline{2500} & 12.60 & 21.27 \\
      10000 & \textbf{12.58} & \textbf{21.24} \\
      \bottomrule
    \end{tabular}
  \end{minipage}
  \hfill % 撑开中间的间距
  \begin{minipage}{0.48\textwidth}
    \centering
    \caption{
    \textbf{Effect of Learning Rate.} The optimal learning rate varies by model architecture. Both overly small and excessively large learning rates lead to suboptimal convergence (WikiText-2 Word PPL $\downarrow$).}
    \label{tab:lr-tuning}
    \begin{tabular}{ccc}
      \toprule
      Learning Rate & Llama3-1B & Qwen3-1.7B \\
      \midrule
      5e-5 & 12.78 & 21.33 \\
      1e-4 & 12.69 & \textbf{21.27} \\
      5e-4 & \textbf{12.60} & 21.45 \\
      1e-3 & 12.82 & 21.91 \\
      \bottomrule
    \end{tabular}
  \end{minipage}
\end{table}
\paragraph{Hyperparameter Sensitivity Analysis} 
We further investigate the robustness of our method by analyzing the influence of key training hyperparameters during the alignment.

As illustrated in Table~\ref{tab:steps-tuning}, increasing the number of optimization steps consistently improves model performance, as it allows the rounding parameters to better adapt to the global activation distribution. However, we observe significant diminishing marginal returns after 2,500 steps. For instance, scaling the training from 2,500 to 10,000 steps (a 4$\times$ increase in computational cost) yields only marginal perplexity reductions of 0.02 for Llama3-1B (12.60 to 12.58) and 0.03 for Qwen3-1.7B (21.27 to 21.24). Therefore, we select 2,500 steps as the default setting to strike an optimal balance between computational efficiency and quantization quality.
    
Table~\ref{tab:lr-tuning} highlights the sensitivity of the rounding optimization to the learning rate, revealing different optimal learning rates across architectures. Specifically, Llama3-1B achieves its optimal perplexity of 12.60 at a learning rate of 5e-4, whereas Qwen3-1.7B reaches its best performance of 21.27 at 1e-4. This discrepancy suggests that the optimal step size depends on the specific weight variance and loss landscape of the respective models. Overly small learning rates lead to suboptimal convergence, while excessively large values degrade performance, likely due to training instability and overshooting the optimal rounding parameters.

\section{Conclusion}
\label{conclusion}
In this work, we present a learnable NVFP4 quantization framework that synergistically integrates FAAR with a 2FA fine-tuning scheme. By addressing the fundamental mismatch between the non-uniform NVFP4 numerical grid and conventional training-free rounding assumptions, our method reformulates rounding as a learnable optimization problem, effectively approximating the theoretically optimal quantization for both weights and activations. Extensive experiments on the Llama3 and Qwen3 model families demonstrate that our method consistently outperforms state-of-the-art approaches across various benchmarks and zero-shot downstream tasks. Remarkably, these significant performance gains incur a minimal training overhead of only 4 GPU hours on 1B-parameter models. We hope this work establishes the necessity of format-aware optimization for irregular numerical formats and facilitates the practical deployment of NVFP4-based large language models on resource-constrained edge devices.

\bibliographystyle{unsrt}
\bibliography{refs}

\end{document}